\documentclass[runningheads]{llncs}

\usepackage{eccv}

\usepackage{eccvabbrv}

\usepackage{graphicx}
\usepackage{booktabs}
\usepackage{multirow}
\usepackage[table]{xcolor}
\usepackage{listings}
\usepackage{subcaption}
\usepackage{tcolorbox}
\tcbuselibrary{listings, breakable, skins}

\newtcblisting{promptbox}[1]{
    enhanced,
    breakable,              
    colback=gray!5,         
    colframe=gray!40,       
    boxrule=0.5pt,          
    arc=3pt,                
    left=4pt, right=4pt, top=2pt, bottom=2pt,
    title=#1,               
    fonttitle=\bfseries\small, 
    coltitle=black,         
    titlerule=0.5pt,
    toptitle=3pt, bottomtitle=3pt,
    listing only,
    listing options={
        basicstyle=\small, 
        breaklines=true,                 
        breakatwhitespace=true,
        breakindent=0pt,                 
        columns=fullflexible,            
        keepspaces=true,
        showstringspaces=false
    }
}

\usepackage[accsupp]{axessibility}  

\usepackage{hyperref}

\usepackage{orcidlink}

\newcommand{\model}[0]{\text{TennisExpert}}
\newcommand{\dataset}[0]{\text{TennisVL}}

\usepackage{amssymb}
\usepackage{pifont}
\newcommand{\cmark}{\ding{51}}%
\newcommand{\xmark}{\ding{55}}%

\begin{document}

\title{Towards Automated Sports Video Understanding: A Tactic-Aware Multimodal Framework for Expert Commentary Generation}
\title{From Perception to Tactical Reasoning: Towards Analytical Sports Commentary Generation}
\title{\model{}: Towards Comprehensive Sports Video Understanding and Analytical Commentary Generation}
\title{\model{}: Towards Expert-Level Analytical Sports Video Understanding}

\titlerunning{\model{}: Towards Expert-Level Analytical Sports Video Understanding}

\author{
Zhaoyu Liu\inst{1} \and
Xi Weng\inst{1} \and
Lianyu Hu\inst{2}\thanks{Corresponding author: hulianyuyy@gmail.com} \and 
Zhe Hou\inst{3} \and
Kan Jiang\inst{1} \and
Jin Song Dong\inst{1} \and
Yang Liu\inst{2}
}

\authorrunning{Z.~Liu et al.}

\institute{National University of Singapore \and
Nanyang Technological University \and 
Griffith University
}

\maketitle

\begin{abstract}

  Tennis is one of the most widely followed sports, generating extensive broadcast footage with strong potential for professional analysis, automated coaching, and real-time commentary. However, automatic tennis understanding remains underexplored due to two key challenges: (1) the lack of large-scale benchmarks with fine-grained annotations and expert-level commentary, and (2) the difficulty of building accurate yet efficient multimodal systems suitable for real-time deployment. 
  To address these challenges, we introduce \dataset{}, a large-scale tennis benchmark comprising over 200 professional matches (471.9 hours) and 40,000+ rally-level clips.
  Unlike existing commentary datasets that focus on descriptive play-by-play narration, \dataset{} emphasizes expert analytical commentary capturing tactical reasoning, player decisions, and match momentum. Furthermore, we propose \model{}, a multimodal tennis understanding framework that integrates a video semantic parser with a memory-augmented model built on Qwen3-VL-8B. The parser extracts key match elements (e.g., scores, shot sequences, ball bounces, and player locations), while hierarchical memory modules capture both short- and long-term temporal context. Experiments show that \model{} consistently outperforms strong proprietary baselines, including GPT-5, Gemini, and Claude, and demonstrates improved ability to capture tactical context and match dynamics. 
  Our dataset and code are publicly available at \url{https://github.com/LZYAndy/TennisExpert}.

\end{abstract}

\section{Introduction}
\label{sec:intro}

\begin{figure}[t]
  \centering
  \setlength{\abovecaptionskip}{5pt}
  \includegraphics[width=\linewidth]{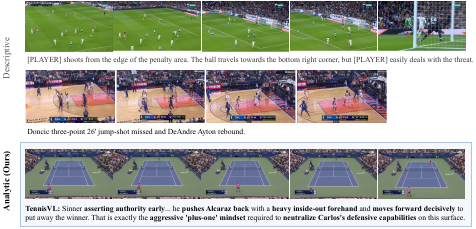} 
  \caption{Existing sports commentary datasets primarily provide descriptive narration of immediate actions and visual states (top and middle rows). In contrast, TennisVL provides expert-level analytical commentary (bottom row), emphasizing tactical intent, match momentum, and performance evaluation beyond surface-level descriptions.}
  \label{fig:commentary_sports}
  \vspace{-20px}
\end{figure}

Tennis is one of the most widely followed sports, attracting millions of audiences across major tournaments such as the Australian Open, Roland Garros, Wimbledon, and the US Open. Each year, thousands of hours of professional match footage are recorded and broadcast, creating a vast amount of visual data with strong potential for professional analysis, automated coaching, and real-time commentary.
General video understanding tasks such as video captioning~\cite{maaz2023video,wang2024internvideo2,lian2025describe} and action recognition~\cite{alomar2025cnns,qian2025dispider,yu2025learning} have achieved remarkable progress. Sports analytics in domains such as soccer~\cite{gautam2024soccernet,you2025timesoccer} and basketball~\cite{wu2022sports,li2025machine} have also benefited from systematic benchmarks and analysis. However, automatic tennis understanding remains largely unexplored due to its unique characteristics~\cite{o2001notational,palut2005dynamical}. 
Tennis analysis requires fine-grained temporal perception to capture fast-moving balls and rapid shot exchanges. It also requires long-range contextual reasoning over match dynamics that can span hours. Moreover, unlike soccer or basketball where commentary accompanies continuous play, tennis commentary is delivered after each rally during brief intervals between points, requiring higher-level analysis beyond surface-level narration as shown in Figure~\ref{fig:commentary_sports}.


Two fundamental challenges hinder progress in automatic tennis understanding.
\textbf{First}, existing tennis datasets~\cite{faulkner2017tenniset,yan2016generating} are limited in scale, diversity, and annotation quality. Most datasets contain short clips from a small number of matches and provide only coarse annotations, such as shot types or scores. They lack diverse match scenarios, fine-grained temporal annotations, and expert-level narration. As a result, current datasets cannot fully capture the complexity of professional tennis. Without a large-scale benchmark covering serve dynamics, rally structures, and tactical patterns, developing and fairly evaluating tennis analysis systems remains difficult.
\textbf{Second}, building a multimodal analysis system that is both accurate and efficient for real-time deployment is technically challenging. Tennis is a fast-paced sport where balls can exceed speeds of 200 km/h and rallies evolve within fractions of a second. Understanding a single point requires jointly reasoning about player positions, ball trajectories, and landing locations. An effective system must detect these fine-grained visual cues in real time while maintaining contextual awareness across rallies and games to generate coherent and informative commentary.



To address these challenges, we first introduce \textbf{\dataset{}}, a \textbf{large-scale tennis benchmark} built from 202 broadcast matches (471.9 hours) and segmented into 40,523 rally-level clips. Each clip is annotated with high-quality natural language captions with expert-level analysis, enabling research on real-time commentary and fine-grained tennis understanding. To the best of our knowledge, this is the largest and most richly annotated tennis video benchmark to date.
Building upon this benchmark, we also propose \textbf{\model{}}, a comprehensive tennis understanding system with two tightly integrated components. 
The first is a \textbf{fine-grained video semantic parser} that parses raw video into compact, time-aligned semantic representations in real time, including scoreboard states, shot sequences, ball bounces, and player locations.
The second component is a \textbf{memory-augmented multimodal model} based on Qwen3-VL-8B~\cite{bai2025qwen3vl}. 
It incorporates a short-term memory module to model recent rallies and a long-term memory module to retain contextual information across games and sets. 
Unlike approaches that directly transcribe long raw videos into commentary with high computational cost and low information density, our dual-memory architecture enables efficient reasoning over both instantaneous events and evolving match dynamics. 
Extensive experiments show that \model{}, despite its compact 8B size, significantly outperforms both open-source models such as InternVL and the Qwen3-VL series, as well as substantially larger proprietary models including GPT-5, Gemini, and Claude. 
We further provide comprehensive visual analyses that illustrate the model's capability for professional-level tennis understanding under complex match scenarios.

In summary, our contributions are as follows:
\begin{itemize}
  \item We construct \dataset{}, the largest tennis video benchmark to date, comprising over 200 match videos with more than 40,000 annotated clips, each paired with high-quality captions for real-time commentary.
  \item We propose \model{}, a comprehensive tennis understanding framework that integrates a video semantic parser with a memory-augmented multimodal model to bridge raw visual inputs and structured tennis analysis.
  \item Extensive experiments show that \model{} significantly outperforms all baselines and enables expert-level tennis understanding in real time.
\end{itemize}

\section{Related Works}
\subsection{Multimodal Models}

Early multimodal large language models (MLLMs)~\cite{li2023blip2,girdhar2023imagebind,guo2024llava,liu2024llava,li2024mini}, such as Flamingo~\cite{alayrac2022flamingo}, BLIP-2~\cite{li2023blip2}, and LLaVA~\cite{liu2024llava}, established vision–language integration by linking visual encoders with large language models via cross-attention mechanisms, learnable query tokens, and visual instruction tuning.

The field has since progressed rapidly, with increasingly powerful architectures emerging~\cite{zhu2025internvl3,bai2025qwen3vl,lu2025ovis2,team2025kimivl}. The InternVL series~\cite{chen2024internvl25, zhu2025internvl3} introduces native multimodal pre-training and enhances post-training strategies such as test-time scaling and tool integration, achieving leading results across diverse benchmarks. The Qwen-VL series~\cite{bai2025qwen25vl,bai2025qwen3vl} has advanced swiftly, culminating in Qwen3-VL~\cite{bai2025qwen3vl}, which attains state-of-the-art performance through architectural innovations. Kimi-VL~\cite{team2025kimivl} adopts a mixture-of-experts design optimized for efficient long-context processing. Ovis2~\cite{lu2025ovis2} improves visual embedding through hierarchical visual token generation, strengthening visual representation learning.

Proprietary systems have also made substantial strides. GPT-5~\cite{singh2025openai} showcases native multimodal reasoning with notable gains in speed and capability. Claude-4.1-Opus~\cite{claude35sonnet} demonstrates strong visual reasoning and improved instruction-following abilities. Gemini~\cite{comanici2025gemini} further advances multimodal performance by incorporating native tool use and agentic functionalities.

Despite these advances, existing frontier MLLMs still struggle with fine-grained, domain-specific professional sports analysis, such as high-level tennis strategy interpretation, where nuanced temporal dynamics, tactical intent, and contextual match understanding remain challenging.

\vspace{-5px}
\subsection{Sports Understanding and Analysis}
\vspace{-2px}
Sports video understanding has become a key research area, driven by applications in automated commentary, tactical analysis, and coaching. Early works in this field focused on event detection and action recognition~\cite{giancola2018soccernet,shao2020finegym,xu2022finediving,liu2025f}.
However, these efforts typically focus on isolated events and lack the ability to analyze long-term game strategies or momentum shifts, which are crucial for comprehensive sports understanding.

Most existing sports video captioning methods and datasets~\cite{faulkner2017tenniset,wu2022sports,mkhallati2023soccernet,ge2024scbench,rao2024matchtime,li2025multi,you2025timesoccer} focus on descriptive captions that summarize immediate actions (e.g., player movements or visual states). They typically rely on sparse frame sampling to balance long temporal spans and computational cost, inevitably discarding fine-grained details in fast-paced exchanges. Moreover, they lack mechanisms to preserve long-term context. As a result, while effective for short summaries, they struggle with deeper analysis such as tactical reasoning, performance evaluation, and momentum tracking, which are essential for expert-level commentary.

To address the gap, we introduce TennisVL, a large-scale benchmark for comprehensive tennis match analysis. Unlike prior datasets, TennisVL emphasizes analytical commentary with expert insights into strategy, match dynamics, and momentum shifts. By integrating structured metadata and expert-level narration, it bridges low-level event/object detection and high-level tactical reasoning, enabling more advanced sports understanding systems.
\vspace{-5px}
\section{Dataset}
\vspace{-2px}

Existing sports commentary datasets primarily provide descriptive play-by-play narration of observable actions \cite{wu2022sports,mkhallati2023soccernet,rao2024matchtime,faulkner2017tenniset}, with limited analytical depth. To address this, we introduce \dataset{}, a large-scale tennis benchmark that combines expert-level analytical commentary with rich multimodal metadata (e.g., audio transcripts, shot sequences, and scoreboard states). Unlike prior datasets, \dataset{} emphasizes tactical intent, match momentum, and performance evaluation beyond surface-level descriptions. Although commentary generation is our primary task, its fine-grained annotations also support broader applications such as automated coaching and action prediction.


\subsection{Dataset Construction}
\subsubsection{Video Collection.} 
We collected 202 publicly available, high-resolution ($1280 \times 720$, 25--30 FPS) singles tennis matches between 2019 and 2025 from YouTube. To ensure comprehensive representation, the collection spans all four Grand Slam tournaments (Australian Open, Roland Garros, Wimbledon, and US Open), capturing varied court surfaces (hard, clay, and grass), diverse professional players (both ATP and WTA), as well as right- and left-handed competitors.

\vspace{-10px}
\subsubsection{Automated Rally Segmentation.}
Unlike soccer or basketball, where commentary is delivered concurrently with play, professional tennis commentary typically occurs \textit{after} each rally, during the interval \textit{between} points, and emphasizes high-level analysis. Accordingly, we organize our dataset at the rally level.
To efficiently process hour-long broadcasts, we train a lightweight audio neural network to detect the distinctive acoustic signatures of racquet–ball impacts. Temporally localized, high-confidence impact events are clustered to segment full matches into candidate rally clips. We then apply filtering to retain only clips that (1) exhibit standard broadcast top-down court views and (2) contain visible scoreboards, ensuring that each segment corresponds to a single, uninterrupted rally. Additional details are provided in the Appendix.

\vspace{-10px}
\subsubsection{Commentary Annotation Pipeline.}
A common approach to obtaining commentary ground truth is to extract transcripts directly from broadcast audio~\cite{gautam2024soccernet,ge2024scbench}. However, raw transcripts are often noisy and inconsistent, reflecting varying expertise, off-topic discussions, cross-talk, and limited tactical depth, which makes them unsuitable as reliable ground truth for analytical commentary.
To construct consistent and high-quality rally-level references, we design an automated multimodal fusion pipeline. First, broadcast audio is transcribed using WhisperX~\cite{bain2023whisperx} and temporally aligned with rally clips. Second, we incorporate structured, human-annotated shot-by-shot descriptions scraped from TennisAbstract~\cite{tennisabstract}. Finally, an advanced LLM (Gemini 3 Pro) synthesizes match information (e.g., tournament, round), current score, detailed shot sequences, and aligned transcripts into a refined \textit{reference commentary}. To ensure temporal coherence, the model is provided with structured context from preceding rallies, enabling it to capture tactical patterns and momentum shifts. The LLM is prompted to act as an expert analyst and generate concise, engaging, and factually grounded commentary. Detailed prompts are provided in the Appendix.

\begin{table}[t]
\centering
\setlength{\belowcaptionskip}{4pt}
\setlength{\tabcolsep}{1.5mm} 
\renewcommand{\arraystretch}{1.15} 
\caption{
\textbf{Comparison of TennisVL with existing sports video commentary datasets.}
``\#Clips'': annotated video–text pairs; ``Event Seq.'': action sequences;
``ASR Trans.'': aligned audio transcripts; ``Analytic'': analytic/tactical reasoning ($\bigcirc$ = partial).
}
\label{tab:dataset_comparison}
\resizebox{\textwidth}{!}{%
\begin{tabular}{l l c c c c c}
\toprule
\textbf{Dataset} & \textbf{Sport} & \textbf{Video (h)} & \textbf{\# Clips} & \textbf{Event Seq.} & \textbf{ASR Trans.} & \textbf{Analytic} \\
\midrule
SoccerNet-Caption~\cite{mkhallati2023soccernet} & Soccer     & 715.9 & 36,894   & \cmark   & \xmark & \xmark \\
MatchTime~\cite{rao2024matchtime}         & Soccer     & 715.9 & 32,743   & \cmark   & \cmark & \xmark \\
GOAL~\cite{qi2023goal}              & Soccer     & 25.0+ & 8,981    & \xmark   & \xmark & $\bigcirc$   \\
SVN~\cite{yan2019fine}               & Basketball & 7.7   & 9,632    & \xmark   & \xmark & \xmark \\
NSVA~\cite{wu2022sports}              & Basketball & 84.8  & 32,019   & \cmark & \xmark & \xmark \\
Tennis Dataset~\cite{yan2016generating}    & Tennis     & 1.5   & 633      & \xmark   & \xmark & \xmark \\
TenniSet~\cite{faulkner2017tenniset}          & Tennis     & N/A   & 746      & \cmark & \xmark & \xmark \\
SCBench~\cite{ge2024scbench}           & Multiple   & N/A   & 5,775    & \xmark   & \cmark & $\bigcirc$   \\
\midrule
\rowcolor{gray!15} \textbf{TennisVL (Ours)}& \textbf{Tennis} & \textbf{471.9} & \textbf{40,523} & \textbf{\cmark} & \textbf{\cmark} & \textbf{\cmark} \\
\bottomrule
\end{tabular}%
}
\vspace{-10px}
\end{table}

\vspace{-10px}
\subsubsection{Annotation Verification.}
To mitigate LLM hallucinations and ensure factual accuracy, we employ a two-stage verification protocol. In the first stage, an \textit{automated sanity check} deterministically ensures that critical entities in the generated commentary (e.g., player names, current scores, and shot types) strictly correspond to the reference metadata (i.e., ground-truth shot sequences, audio transcripts, and match statistics). Outputs failing this consistency check are automatically rejected and regenerated. In the second stage, we perform a \textit{human expert review} on a random subset of 2,000 video-commentary pairs. Domain experts assess the annotations based on factual alignment, tactical correctness, and appropriate use of professional terminology, with their feedback used to iteratively refine the generation prompts. Ultimately, the human acceptance rate exceeded 95\%. Any remaining substandard annotations were manually corrected or discarded, establishing \textit{TennisVL} as a robust and reliable benchmark.

\subsection{Data Structure} \label{data_structure}
Our dataset is structured as a collection of JSON objects, where each entry represents a single rally and its associated multi-modal metadata. The schema is designed to easily facilitate downstream tasks in video understanding and natural language generation:


\begin{lstlisting}[basicstyle=\ttfamily\scriptsize]
[
  {
    "clip_id": "matchID_start_end", // Match ID with temporal boundaries
    "match_info": {...}, // Tournament, round, surface, player info
    "scoreboard": {
        "Player A": [1, 2, 30], // Score: Set, Game, Point
        "Player B": [0, 3, 15], 
        "server": "Player A"
    },
    "audio_transcript": "...", // Temporally aligned WhisperX output
    "shot_sequence": [ ... ], // Structured shot-by-shot breakdown 
    "commentary": "..." // Final LLM-synthesized reference commentary
  },
  ...
]
\end{lstlisting}


Beyond commentary generation, this comprehensive schema inherently supports a variety of advanced sports understanding tasks. For instance, the integration of temporally aligned \texttt{shot\_sequence} and \texttt{scoreboard} data provides ideal ground truth for automated coaching, professional tactical analysis, and fine-grained action prediction, which we leave for future exploration.




\subsection{Dataset Statistics}
Table~\ref{tab:dataset_comparison} compares TennisVL with existing sports video commentary datasets. The key strengths of TennisVL are its scale, comprehensive multimodal alignment, and analytical depth. With 400+ hours of broadcast video and over 40k annotated clips, it substantially exceeds prior tennis-specific datasets and is competitive with the largest sports benchmarks. Constructed from publicly available broadcasts through a semi-automated pipeline with human verification, the dataset is scalable and can be readily extended as new matches become available.

\begin{figure}[t]
    \centering
    \setlength{\abovecaptionskip}{3pt}
    \begin{subfigure}[t]{0.30\linewidth}
        \centering
        \includegraphics[width=\linewidth]{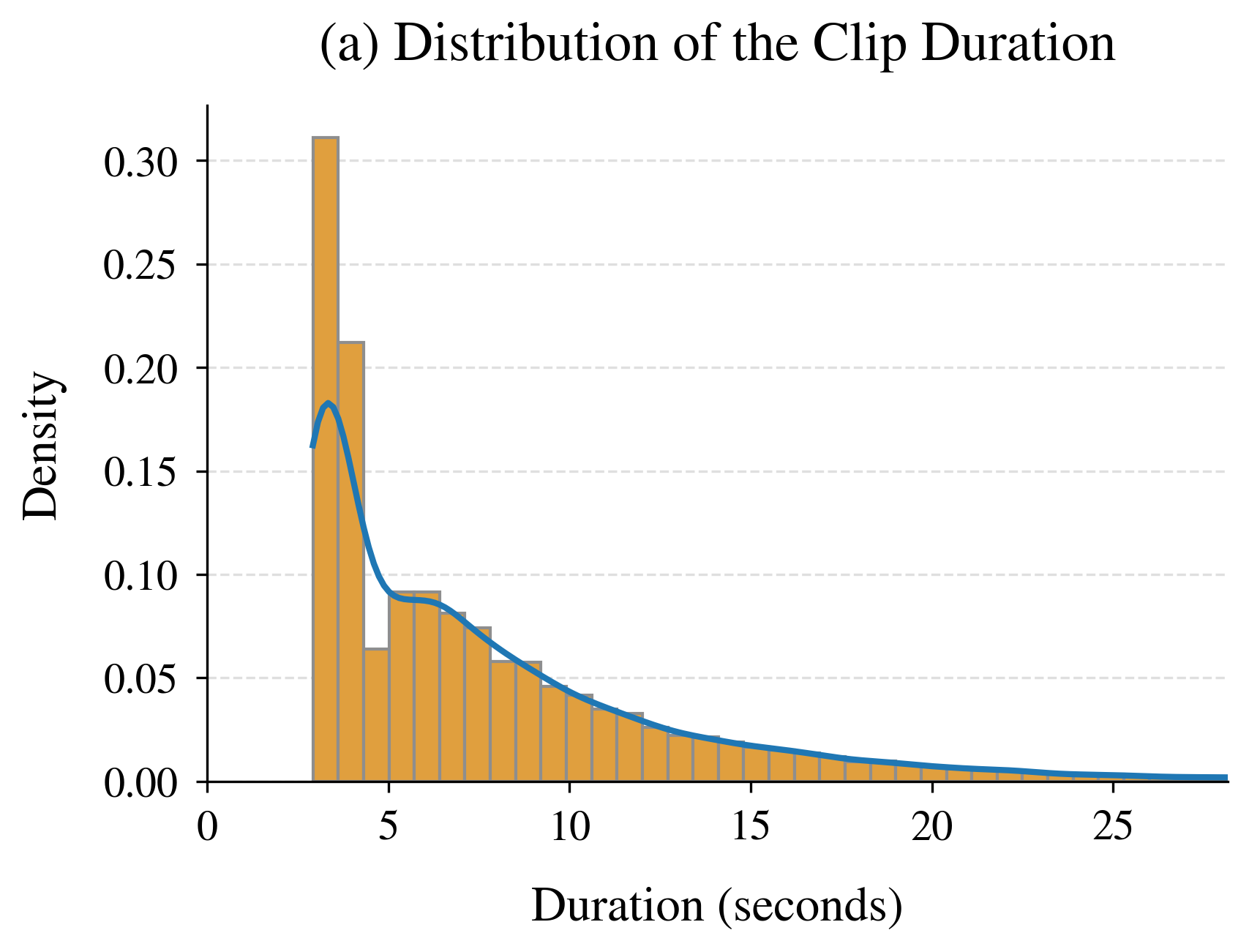}
    \end{subfigure}
    \hspace{-0.00\linewidth}
    \begin{subfigure}[t]{0.30\linewidth}
        \centering
        \includegraphics[width=\linewidth]{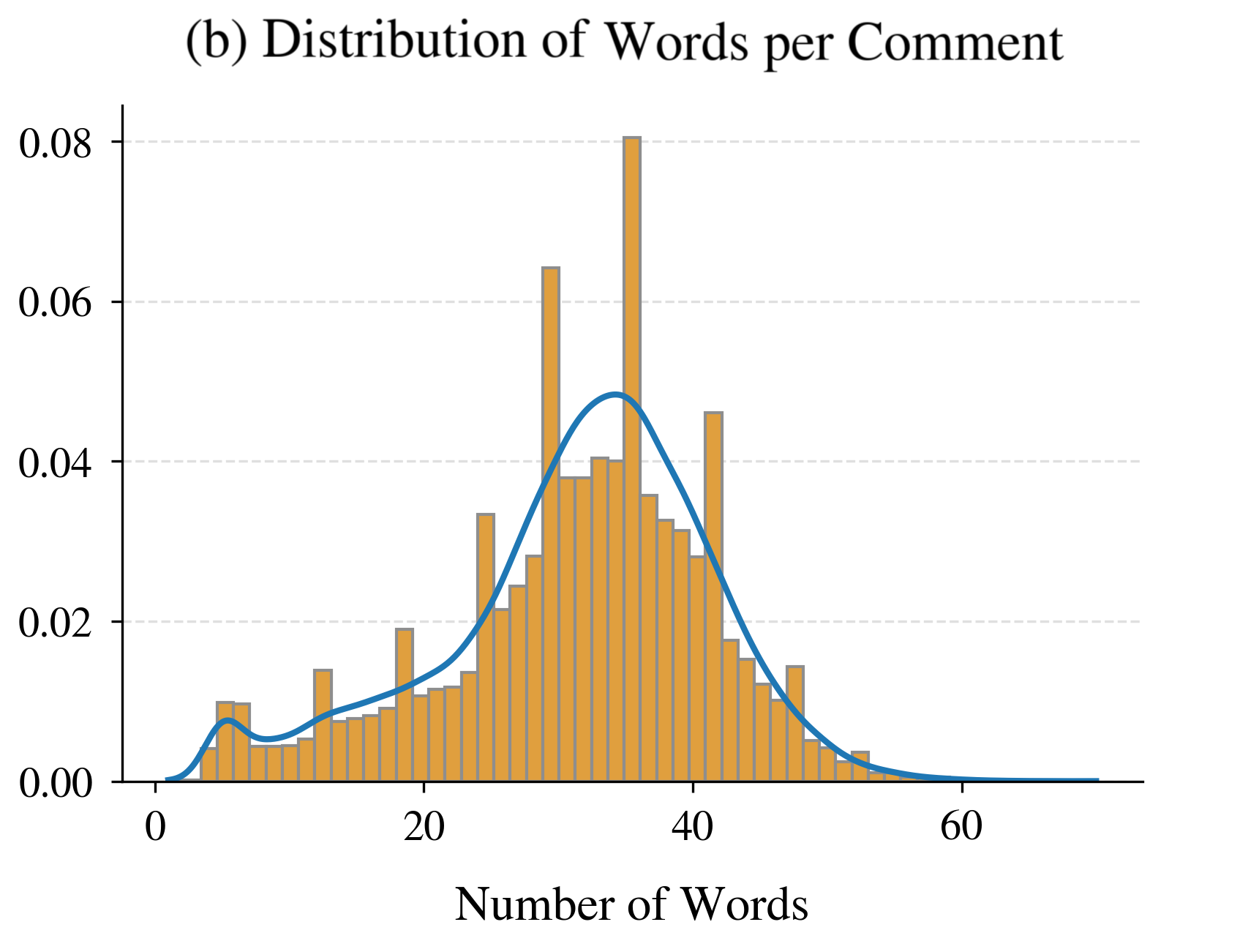}
    \end{subfigure}
    \hspace{-0.042\linewidth}
    \begin{subfigure}[t]{0.40\linewidth}
        \centering
        \includegraphics[width=\linewidth]{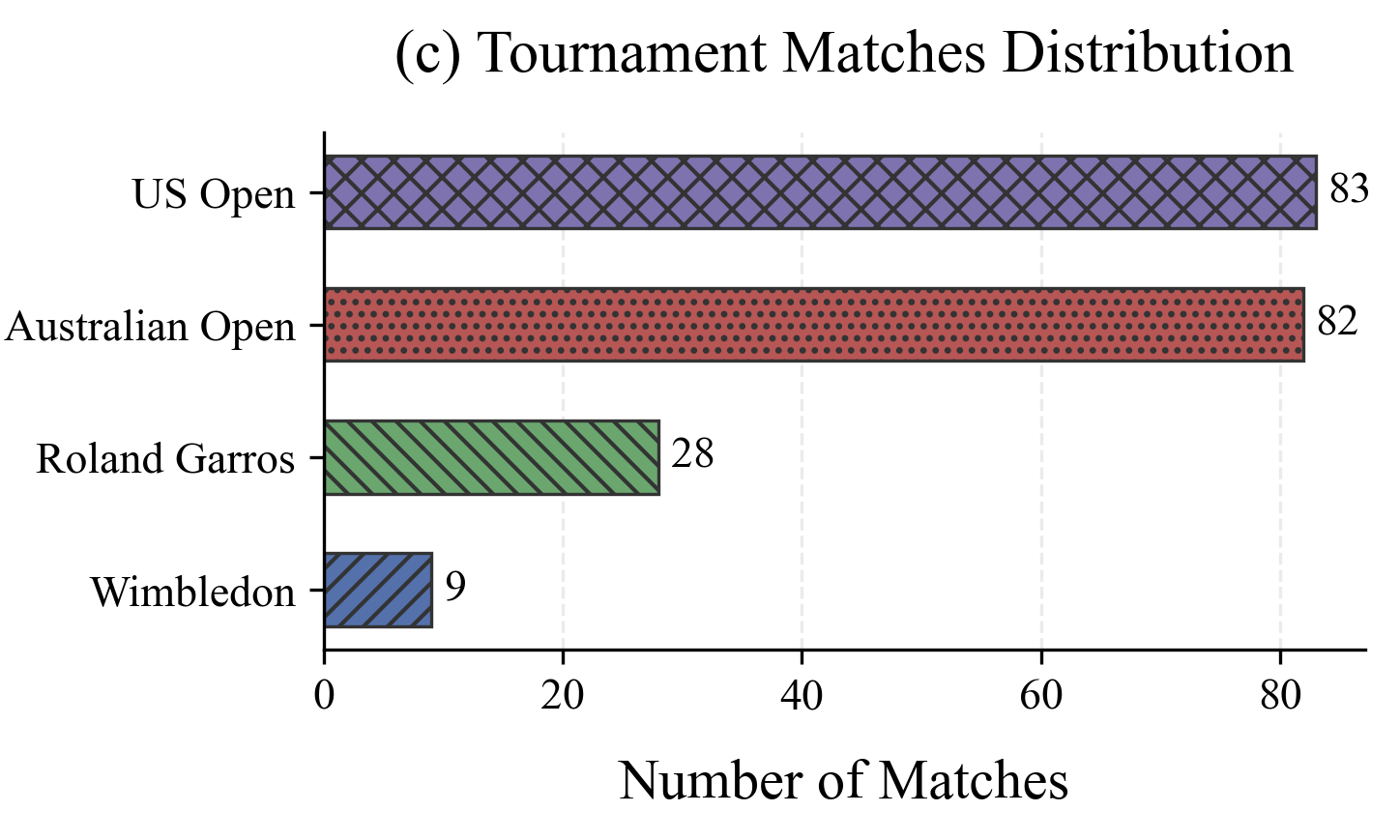}
    \end{subfigure}
    
    \caption{
    Dataset statistics of clip duration, commentary length, and tournaments.
    }
    \label{fig:dataset_statistics}
    \vspace{-10px}
\end{figure}

TennisVL contains 202 full broadcast matches totaling 471.9 hours of video, segmented into 40,523 rally-level clips with an average duration of 7.68 seconds. The associated expert commentaries average 31.42 words. The dataset covers 94 unique players and 162,503 total shots. The distributions of clip duration, commentary length, and tournament coverage are illustrated in Figure~\ref{fig:dataset_statistics}. 

We split the data at the match level to prevent leakage: 182 matches (35,687 clips) for training and 20 disjoint matches (4,836 clips) for testing, ensuring that all clips and annotations from a given match belong to the same split.

\section{Proposed Approach}

\begin{figure*}[t]
  \centering
  \setlength{\abovecaptionskip}{5pt}
  \includegraphics[width=\textwidth]{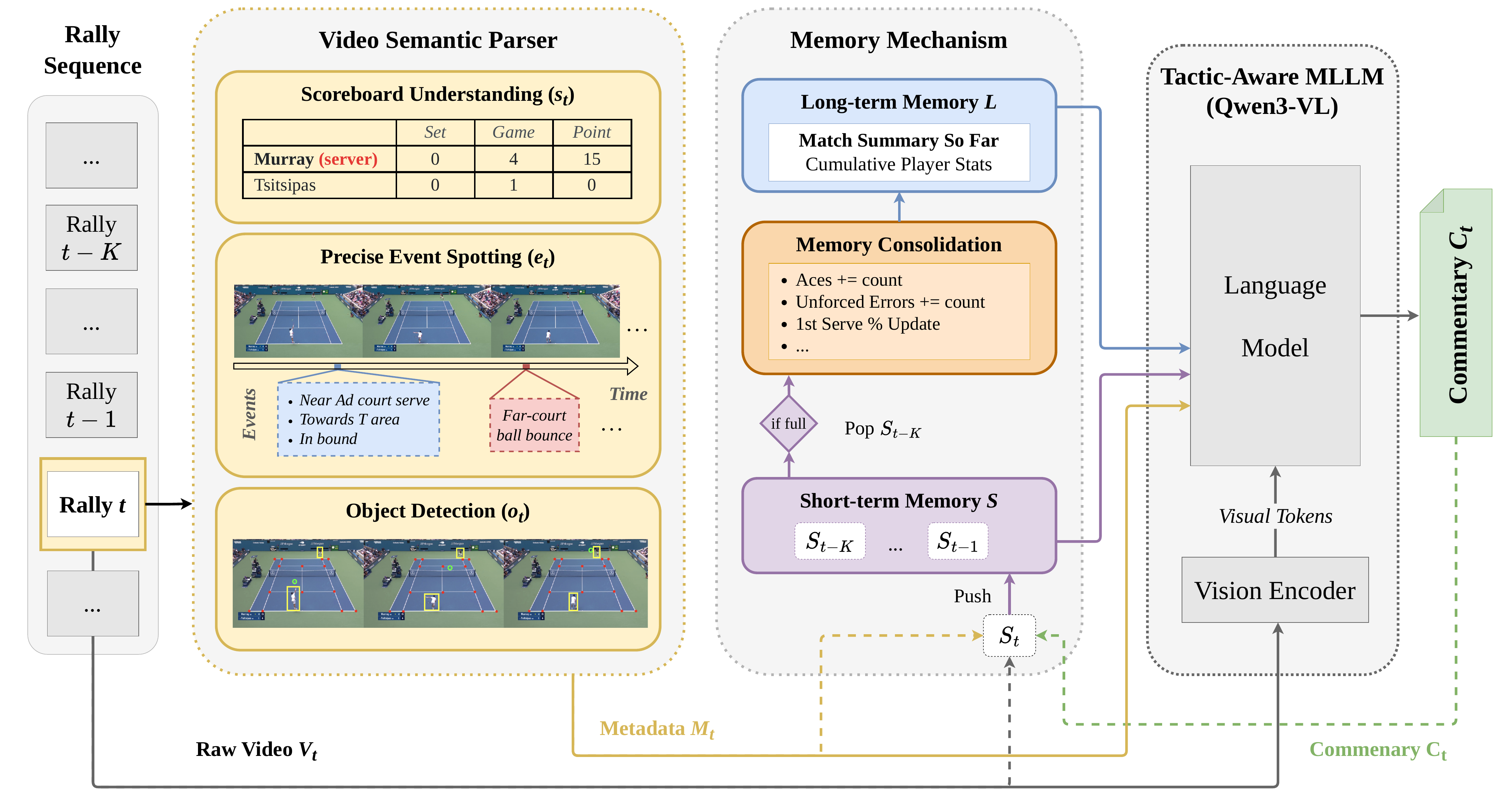} 
  \caption{\textbf{Overall architecture of \model{}.} 
  Given a sequence of rallies, the current rally $V_t$ is processed by a video semantic parser to obtain structured metadata $M_t$, including scoreboard state ($s_t$), fine-grained event sequence ($e_t$), and spatial object detections ($o_t$). A hierarchical memory mechanism maintains match context: short-term memory ($S$) stores recent rally representations in a FIFO buffer, while long-term memory ($L$) consolidates past events into cumulative match statistics. 
  The tactic-aware MLLM (Qwen3-VL) integrates visual tokens, structured metadata, and memory context to generate expert-level commentary $C_t$, which is fed back into short-term memory to preserve match momentum for subsequent rallies.}
  \label{fig:architecture}
  \vspace{-10px}
\end{figure*}



Generating professional tennis commentary requires both fine-grained understanding of sports events (e.g., high-speed action sequences and player–ball dynamics) and long-term context. Existing approaches typically rely on raw video input and need to balance sampling rate with computational cost: sparse sampling may miss critical events, whereas dense sampling is computationally expensive.
To address this, we propose \model{}, a framework that extracts structured semantic metadata and integrates short- and long-term memory mechanisms for precise grounding and historical awareness. For generation, we adopt an MLLM backbone, providing the structured representations as a unified multimodal prompt to produce context-aware, analytical commentary.



\subsection{Overview}

Our framework comprises three components: a video semantic parser, a long short-term memory mechanism, and an MLLM backbone, as illustrated in Figure~\ref{fig:architecture}. Unlike conventional video captioning models that map raw pixels directly to text, our approach supports deeper tactical reasoning by grounding visual signals in structured metadata and historical context.

Given a tennis match represented as a sequence of rallies $\mathcal{S}=\{V_1, V_2, \dots, V_N\}$, where $V_t \in \mathbb{R}^{T \times 3 \times H \times W}$ denotes the $t$-th rally clip with $T$ frames, we first extract structured metadata $M_t$ from $V_t$ via a video analytics pipeline to reduce the reasoning burden on the MLLM. A memory module $\mathcal{H}_t$ aggregates information from preceding rallies to capture match dynamics and momentum. The visual frames, metadata, and memory context are then projected and concatenated as inputs to a tactic-aware MLLM, which autoregressively generates expert-level analytical commentary $C_t$:
\begin{equation}
C_t = \text{MLLM}(V_t, M_t, \mathcal{H}_t).
\end{equation}

\vspace{-10px}
\subsection{Video Semantic Parser}

To bridge raw visual input and high-level tactical reasoning, we extract structured semantic metadata $M_t = \{s_t, e_t, o_t\}$ from each rally clip $V_t$. This representation provides explicit physical and contextual states that are difficult to infer directly from raw visual tokens.

\textbf{Scoreboard understanding ($s_t$).}
We model the scoreboard state as $s_t = (\rho_t^{(A)}, \rho_t^{(B)}, v_t)$, where $\rho_t^{(A)}$ and $\rho_t^{(B)}$ denote the set, game, and point scores of players A and B, and $v_t \in \{A, B\}$ indicates the server. Although MLLMs perform well in OCR, they often struggle with tournament-specific layouts and scoring rules. We therefore adopt a prompt-guided MLLM that incorporates tournament-specific templates and rules to reliably extract the score state. Detailed prompt design and formatting instructions are provided in Appendix.

\textbf{Precise event spotting ($e_t$).}
Extracting a sequence of fine-grained events from fast-paced sports videos remains challenging for existing LLMs. Therefore, we train an end-to-end temporal event detector to extract fine-grained rally events, defined as $e_t = {(c_i, \tau_i)}_{i=1}^{N_e}$, where $c_i \in \mathcal{C}$ is the detailed event classes (e.g., hit events with shot types, techniques, directions, and outcomes, as well as ball-bounce events), $\tau_i$ is the timestamp, and $N_e$ is the number of events. Building on the F$^3$Set dataset~\cite{liu2025f}, which provides dense annotations for detailed hitting events, we \emph{extend the taxonomy with spatial bounce categories} (near- and far-court). The detector combines local spatio-temporal features with GRU to model long-range dependencies, achieving an Edit Score of 81.2 on the test set.

\textbf{Object detection ($o_t$).}
We further perform detection and tracking for players, the ball, and court keypoints across the clip. At each key timestamp $\tau_i$, we sample object coordinates $o_t^{(\tau_i)} = \{\mathbf{b}_t^{(A)}, \mathbf{b}_t^{(B)}, \mathbf{b}t^{(\text{ball})}\}$, where $\mathbf{b} \in \mathbb{R}^2$ denotes the $(x,y)$ center in image coordinates. Court corners are used to estimate a homography matrix, projecting players and ball bounce positions into real-world court coordinates. The aggregated representation $o_t = \bigcup_{i=1}^{N_e} o_t^{(\tau_i)}$ provides a unified structure for geometric reasoning over player positioning and shot placement.




\subsection{Long Short-Term Memory Mechanism}
Professional commentators do not analyze a rally in isolation; they interpret events based on the overarching match momentum and historical player behaviors. Inspired by human cognitive processes, we propose a long short-term memory mechanism $\mathcal{H}_t = [\mathcal{S}_t, \mathcal{L}_t]$.

\textbf{Short-term Working Memory ($\mathcal{S}_t$).} The short-term memory acts as a dynamic buffer that stores the semantic and visual states of the most recent $K$ rallies. It captures the immediate match momentum (e.g., a player winning three consecutive points). It is updated via a First-In-First-Out (FIFO) queue:
\begin{equation}
    \mathcal{S}_t = \{ (V_{i}, M_{i}, C_{i}) \mid i = t-K, \dots, t-1 \}
\end{equation}
where $V_i$, $M_{i}$, $C_{i}$ denote previous clip, metadata, and generated commentary.
\textbf{Long-term Semantic Memory ($\mathcal{L}_t$).}
While the short-term memory tracks transient momentum, the long-term memory serves as a persistent knowledge base summarizing the entire match. Instead of storing raw videos or events, we represent it as a structured state space $\mathcal{L}_t = \{\Omega_t^{(P_1)}, \Omega_t^{(P_2)}\}$, where $\Omega_t^{(p)}$ encodes the cumulative performance statistics of player $p$ up to rally $t$. The memory is incrementally updated via a deterministic consolidation function, $\mathcal{L}_t = \Phi(\mathcal{L}_{t-1}, M_{t-K})$.
When the rally $M_{t-K}$ is removed from the short-term FIFO queue, $\Phi(\cdot)$ parses its structured events to update professional-grade match statistics. This mechanism maintains a comprehensive set of broadcast metrics, including service (e.g., aces, double faults, first-serve percentage), return (e.g., return points won), and overall point statistics (e.g., winners, unforced errors). By compressing historical rallies into dense symbolic summaries, this design enables the agent to address both immediate situational queries and high-level strategic analysis without exceeding the MLLM context window.

\vspace{-3mm}
\subsection{Tactic-Aware Commentary Generation}
For the generation stage, we adopt Qwen3-VL 8B~\cite{bai2025qwen3vl} as the backbone due to its strong visual grounding and instruction-following capabilities. The outputs of preceding modules are tokenized and combined into a unified multimodal prompt. Specifically, the structured metadata $M_t$ and memory context $\mathcal{H}_t$ are serialized into textual tokens $\mathbf{Z}_s^{(t)}$, while the video clip $V_t$ is encoded into visual tokens $\mathbf{Z}_v^{(t)}$ via the model’s visual encoder. A task-specific prompt instructs the model to assume an expert commentator persona.
The commentary $C_t = \{w_1, \dots, w_L\}$ is generated autoregressively by maximizing:
\begin{equation}
    \log p(C_t \mid V_t, M_t, \mathcal{H}_t) = \sum_{j=1}^{L} \log p_{\theta}(w_j \mid w_{<j}, \mathbf{Z}_v^{(t)}, \mathbf{Z}_s^{(t)})
\end{equation}
where $\theta$ denotes the model parameters and $L$ the sequence length. This formulation aligns dense visual signals with structured symbolic context to produce tactically grounded commentary.

\vspace{-9px}
\subsubsection{Inference.}


During the inference phase, \model{} processes a tennis match online. For each incoming rally $V_t$, the semantic pipeline extracts structured metadata $M_t = \{s_t, e_t, o_t\}$ in real-time.
The short- and long-term memory ($\mathcal{S}_t$, $\mathcal{L}_t$) are updated iteratively as the match progresses. When generating $t$-th rally's commentary, the MLLM retrieves the accumulated context $\mathcal{H}_t$ from previous rallies. The metadata $M_t$ and memory context $\mathcal{H}_t$ are unified with the visual tokens of $V_t$ to form the model input, and the MLLM autoregressively produces the expert commentary $C_t$. The $(V_t, M_t, C_t)$ is then inserted into the short-term memory queue, preserving match momentum for the subsequent rally.

\vspace{-2mm}
\section{Experiments}
In this section, we first present implementation details, evaluation metrics, and baseline models, followed by comprehensive quantitative results and ablation studies to evaluate each module’s contribution. We conclude with a qualitative analysis to further illustrate model performance.

\vspace{-3mm}
\subsection{Implementation Details}
For each rally, we extract structured metadata including scoreboard state, fine-grained event sequences, and spatial object trajectories. Score recognition is performed via OCR on localized scoreboard regions, while temporal event spotting follows an end-to-end dense prediction framework~\cite{liu2025f}. Players and the ball are detected and tracked using a tennis fine-tuned RF-DETR~\cite{sapkota2025rf}. Court keypoints are predicted using a heatmap-based model, and a homography matrix is estimated to project player positions and ball landing points into real-world court coordinates. To model short-term momentum, we adopt a sliding-window memory with $K=4$, retaining the current rally and the previous four rallies. For efficiency, video clips are pre-encoded once per training step and cached across overlapping windows, and loss is computed independently per window to reduce GPU memory usage. We fine-tune \textit{Qwen3-VL-8B-Instruct} via supervised fine-tuning, updating the language model while freezing the vision encoder and multimodal projector. Training runs for 3 epochs with a learning rate of $1\times10^{-5}$ using cosine scheduling on 4 NVIDIA H200 GPUs under BF16 precision. Additional implementation details are provided in the Appendix.

\vspace{-3mm}
\subsection{Evaluation Metrics}

To evaluate commentary quality, we employ both standard natural language generation metrics and a domain-specific LLM-based evaluation protocol.

\vspace{-13px}
\subsubsection{Standard Metrics.}
We report various automatic evaluation metrics for text generation, including BLEU-4~\cite{papineni2002bleu}, METEOR~\cite{banerjee2005meteor}, ROUGE-L~\cite{lin2004rouge}, and CIDEr~\cite{vedantam2015cider}.

\vspace{-13px}
\subsubsection{Expert LLM Score.}
Since standard metrics do not fully capture tactical depth or factual accuracy, we introduce an LLM-based scoring framework. 
An LLM evaluator acts as a senior tennis analyst and assesses the predicted commentary against the ground-truth annotations described in Section~\ref{data_structure}. It assigns a score from 0 to 100 based on five equally weighted criteria (20 points each):
\vspace{-3px}


\begin{enumerate}
\item \textbf{Accuracy:} Factual alignment with the ground truth data, including correct player identities, shot types, score states, and court positions. 

\item \textbf{Coherence:} Logical narrative flow and consistent referential usage. 

\item \textbf{Excitement:} Appropriateness of tone relative to rally intensity. 

\item \textbf{Professionalism:} Depth of tactical insight and correct use of domain-specific terminology. (e.g., shot patterns, strategic adjustments).

\item \textbf{Pacing:} Length and level of detail relative to rally complexity. Concise for simple points and more descriptive for extended, strategically rich rallies.
\end{enumerate}
\vspace{-3px}

The full prompt and formatting instructions are provided in Appendix.

\vspace{-3mm}
\subsection{Baselines}
We compare our framework with leading MLLMs, categorized into open-source and proprietary models. The open-weight baselines include InternVL 3.5~\cite{wang2025internvl3}, Qwen3-VL~\cite{bai2025qwen3vl}, and Qwen3.5~\cite{qwen35blog}. The proprietary baselines comprise Claude 4.5 Sonnet~\cite{claude35sonnet}, Gemini 3 Flash \& Pro~\cite{comanici2025gemini}, and GPT-5.2~\cite{singh2025openai}. For a fair comparison, all models are provided with identical raw video inputs and prompted to generate professional tennis commentary under the same instructions.

\subsection{Quantitative Results}

\begin{table}[t]
\centering
\setlength{\belowcaptionskip}{3pt}
\setlength{\tabcolsep}{1mm} 
\renewcommand{\arraystretch}{1.15} 
\caption{Quantitative evaluation results on traditional metrics and LLM-based metrics. ``Acc'', ``Coh'', ``Exc'', ``Pro'', and ``Pac'' denote ``Accuracy'', ``Coherence'', ``Excitement'', ``Professionalism'', and ``Pacing''. Our \model{} achieves SOTA performance.}
\label{tab:evaluation_results}
\resizebox{\textwidth}{!}{
\begin{tabular}{l cccc cccccc}
\toprule
\multirow{2}{*}{\textbf{Model}} & \multicolumn{4}{c}{\textbf{Traditional Metrics}} & \multicolumn{6}{c}{\textbf{LLM-based Metrics}} \\
\cmidrule(lr){2-5} \cmidrule(lr){6-11}
 & {BLEU-4} & {METEOR} & {ROUGE-L} & {CIDEr} & {Acc} & {Coh} & {Exc} & {Pro} & {Pac} & {Total} \\
\midrule
\multicolumn{11}{l}{\textit{Open-Weight Models (Zero-Shot)}} \vspace{1mm}\\
InternVL3.5-8B      & 0.75 & 20.20 & 13.61 & 0.08  & 0.90  & 5.10  & 6.69  & 2.86  & 3.47  & 19.02 \\
InternVL3.5-38B     & 0.84 & 20.58 & 13.22 & 0.11  & 1.45  & 8.98  & 10.56 & 4.16  & 5.25  & 30.40 \\
Qwen3-VL-8B         & 1.20 & 18.90 & 14.90 & 3.89  & 1.70  & 10.56 & 11.78 & 5.05  & 8.59  & 37.68 \\
Qwen3-VL-32B        & 1.24 & 20.42 & 15.46 & 2.61  & 2.19  & 10.74 & 10.81 & 5.50  & 8.74  & 37.97 \\
Qwen3.5-35B         & 1.19 & 19.66 & 14.61 & 3.70  & 3.32  & 14.11 & 10.95 & 7.18  & 9.12  & 44.67 \\
\midrule
\multicolumn{11}{l}{\textit{Proprietary Models (Zero-Shot)}} \vspace{1mm}\\
Claude-4.5-sonnet   & 1.22 & 19.99 & 14.20 & 2.09  & 3.06  & 15.76 & 12.90 & 9.19  & 9.84  & 50.75 \\
Qwen3.5-plus    & 1.51 & 21.74 & 16.61 & 5.44 & 2.70 & 15.21 & 14.01  & 9.49& 11.14 & 52.55 \\
Gemini-3-flash      & 1.57 & 19.11 & 16.02 & 7.31  & 5.31  & 13.11 & 13.08 & 10.19 & 10.81 & 52.50 \\
GPT-5.2             & 1.32 & 18.39 & 15.95 & 7.52  & 3.73  & 16.02 & 14.26 & 10.82 & 11.76 & 56.59 \\
Gemini-3-pro  & 1.78 & 20.02 & 17.41 & 10.11  &  3.96 & 16.50  & 14.54 & 11.63 & 13.25  & 59.89 \\
\midrule
\textbf{TennisExpert (Ours)} & \textbf{7.98} & \textbf{31.54} & \textbf{29.16} & \textbf{43.71} & \textbf{15.79} & \textbf{19.29} & \textbf{16.84} & \textbf{16.73} & \textbf{19.40} & \textbf{88.05} \\
\bottomrule
\end{tabular}
}
\vspace{-10px}
\end{table}

Table~\ref{tab:evaluation_results} reports performance on both traditional and LLM-based metrics.
Among pre-trained models, larger proprietary systems generally outperform open-weight models, reflecting the benefit of scale. Gemini-3-Pro achieves the highest LLM Total score (59.89), 
followed by GPT-5.2 (56.59), Qwen3.5-Plus (52.55), Gemini-3-Flash (52.50), and Claude-4.5-Sonnet (50.75).
However, baseline models obtain low CIDEr ($\leq$ 10.11) and limited Accuracy and Professionalism, indicating insufficient factual grounding and tactical reasoning for tennis commentary. Open-weight models, including the Qwen3-VL and InternVL series, perform substantially worse, highlighting the difficulty of the task without domain adaptation.

Our \model{} significantly outperforms all baselines across metrics. It achieves a CIDEr score of 43.71, over five times higher than the strongest zero-shot model, and a Total LLM score of 88.05. The model leads in Accuracy (15.79), Coherence (19.29), Excitement (16.84), Professionalism (16.73), and Pacing (19.40), demonstrating strong factual reliability and tactical depth. These results validate the effectiveness of structured grounding and hierarchical memory for expert-level tennis commentary generation.
\vspace{-3mm}

\vspace{-10px}
\subsubsection{Efficiency.}
\label{sec:efficiency}
To evaluate real-time deployment feasibility, we analyze the computational efficiency of our framework. The video semantic parser runs in real time on a single GPU at up to 40 FPS, exceeding standard broadcast rates.
We further evaluate MLLM inference efficiency (Table~\ref{tab:computational_efficiency}). Feeding raw video up to time $t$ yields linearly growing token complexity $\mathcal{O}(T)$, causing high latency and memory issues. In contrast, TennisExpert uses structured representations instead of continuous video, maintaining constant context complexity. This design requires only $\sim$20 GB VRAM and achieves \emph{<2 s} latency, enabling strong performance and tactically rich commentary under real-time broadcast constraints.


\begin{table}[t]
\centering
\setlength{\belowcaptionskip}{2pt}
\setlength{\tabcolsep}{1.5mm} 
\renewcommand{\arraystretch}{1.15} 
\caption{Computational efficiency during inference. Our approach significantly reduces VRAM and Latency compared to processing full raw videos.}
\label{tab:computational_efficiency}
\resizebox{\linewidth}{!}{
\begin{tabular}{l l c c c}
\toprule
\textbf{Model Setup} & \textbf{Input} &\textbf{Token Count} & \textbf{VRAM (GB)} & \textbf{Latency (s)} \\
\midrule
Baselines (Dense video) & Full raw video up to $t$ & $\mathcal{O}(T)$  & OOM & High \\
\model{} (Ours) & Structured $V_t + M_t + H_t$ &$\mathcal{O}(1)$ ($\sim$14k) & $\sim$20 GB & {< 2 s} \\
\bottomrule
\end{tabular}
}
\end{table}

\vspace{-5px}
\subsection{Ablation Studies}

\begin{table}[t]
\centering
\setlength{\belowcaptionskip}{2pt}
\setlength{\tabcolsep}{1mm}
\renewcommand{\arraystretch}{1.15}
\caption{Ablation study on input structure and hierarchical memory. $V_t$, $M_t$, $\mathcal{S}_t$, and $\mathcal{L}_t$ denote video, metadata, short-term memory, and long-term memory, respectively.}
\label{tab:ablation}
\resizebox{\textwidth}{!}{
\begin{tabular}{l c cccc cccccc}
\toprule
\multirow{2}{*}{\textbf{Input}} 
& \multirow{2}{*}{\textbf{Memory}} 
& \multicolumn{4}{c}{\textbf{Traditional Metrics}} 
& \multicolumn{6}{c}{\textbf{LLM-based Metrics}} \\
\cmidrule(lr){3-6} \cmidrule(lr){7-12}
& & BLEU-4 & METEOR & ROUGE-L & CIDEr 
& Acc & Coh & Exc & Pro & Pac & Total \\
\midrule

\multicolumn{12}{l}{\textit{Zero-Shot (Qwen3-VL-8B)}} \\

$V_t$ & -- 
& 1.20 & 18.90 & 14.90 & 3.89
& 1.70 & 10.56 & 11.78 & 5.05 & 8.59 & 37.68 \\

$V_t + M_t$ & -- 
& 1.58 & 21.17 & 16.39 & 4.40
& 7.86 & 14.62 & 14.10 & 11.20 & 13.75 & 61.53 \\

\midrule
\multicolumn{12}{l}{\textit{Fine-Tuned (Qwen3-VL-8B)}} \\

$V_t$ & -- 
& 1.95 & 16.30 & 16.15 & 7.86
& 4.10 & 15.05 & 6.63 & 6.49 & 10.46 & 42.74 \\

$V_t + M_t$ & -- 
& 4.27 & 25.61 & 23.80 & 24.15
& 10.84 & 17.07 & 15.14 & 12.46 & 18.23 & 73.74 \\

$V_t + M_t$ & $\mathcal{S}_t$
& 7.88 & \textbf{31.54} & \textbf{29.24} & 41.92
& 15.49 & 18.75 & \textbf{16.92} & 15.63 & 19.17 & 85.97 \\

\textbf{$V_t + M_t$} & \textbf{$\mathcal{S}_t+\mathcal{L}_t$}
& \textbf{7.98} & \textbf{31.54} & 29.16 & \textbf{43.71}
& \textbf{15.79} & \textbf{19.29} & 16.84 & \textbf{16.73} & \textbf{19.40} & \textbf{88.05} \\

\bottomrule
\end{tabular}
}
\vspace{-10px}
\end{table}





All ablations are conducted on the Qwen3-VL-8B backbone (Table~\ref{tab:ablation}) to quantify the contribution of input structure and hierarchical memory.
Using raw video alone ($V_t$) yields limited performance in both zero-shot and fine-tuned settings. In contrast, incorporating structured metadata ($V_t + M_t$) substantially improves results, increasing CIDEr to 24.15 and the Total score to 73.74, with marked gains in Accuracy (4.10 $\rightarrow$ 10.84) and Professionalism (6.49 $\rightarrow$ 12.46). This highlights the importance of semantic grounding for fine-grained tennis analysis.

Adding short-term memory further improves performance, increasing CIDEr by +17.77 and Total score by +12.23, indicating better modeling of rally continuity and short-term momentum. 
Integrating long-term memory yields the best overall performance. 
The full model ($V_t + M_t$ with $\mathcal{S}_t + \mathcal{L}_t$) achieves 43.71 CIDEr and 88.05 Total. 
Compared with short-term memory alone, long-term memory brings consistent gains, improving CIDEr by +1.79 and Total by +2.08, with additional improvements in Accuracy (+0.30), Coherence (+0.54), Professionalism (+1.10), and Pacing (+0.23). 
These results suggest that modeling long-range match context improves the overall quality of generated tennis commentary.

\vspace{-3mm}
\subsection{Qualitative Analysis}
\begin{figure*}[t]
    \centering
    \setlength{\abovecaptionskip}{3pt}
    \includegraphics[width=\textwidth]{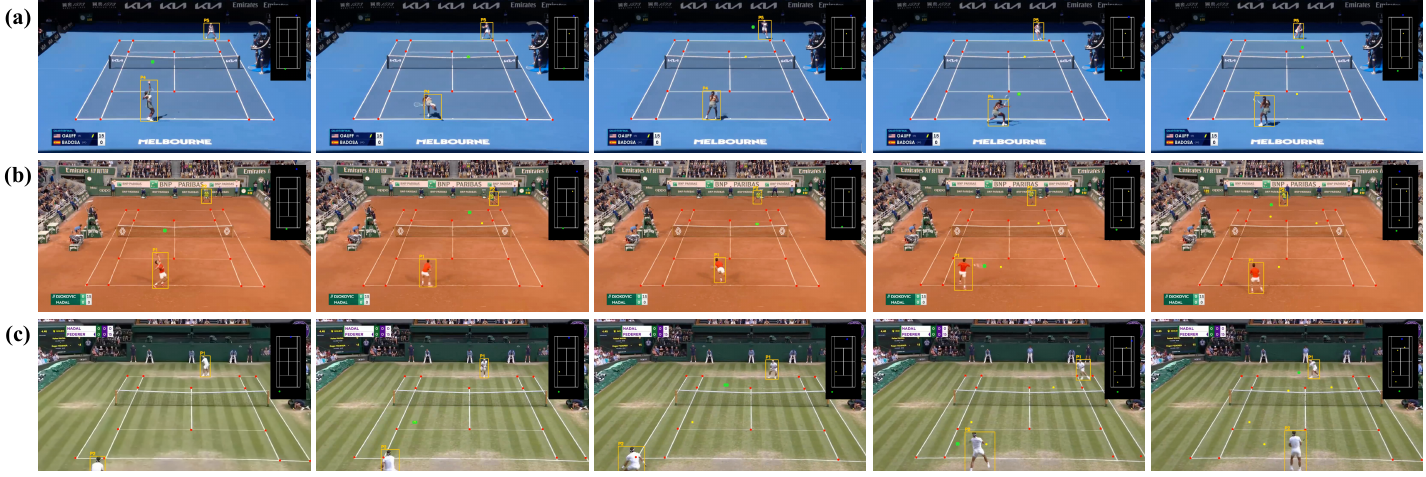}
    \caption{Semantic parser visualization on various court surfaces. Court corners (red), ball (green), bounces (yellow), and player boxes are projected onto broadcast frames.}
    \label{fig:vid_parser}
    \vspace{-7px}
\end{figure*}

\begin{figure*}[t]
    \centering
    \setlength{\abovecaptionskip}{3pt}
    \includegraphics[width=\textwidth]{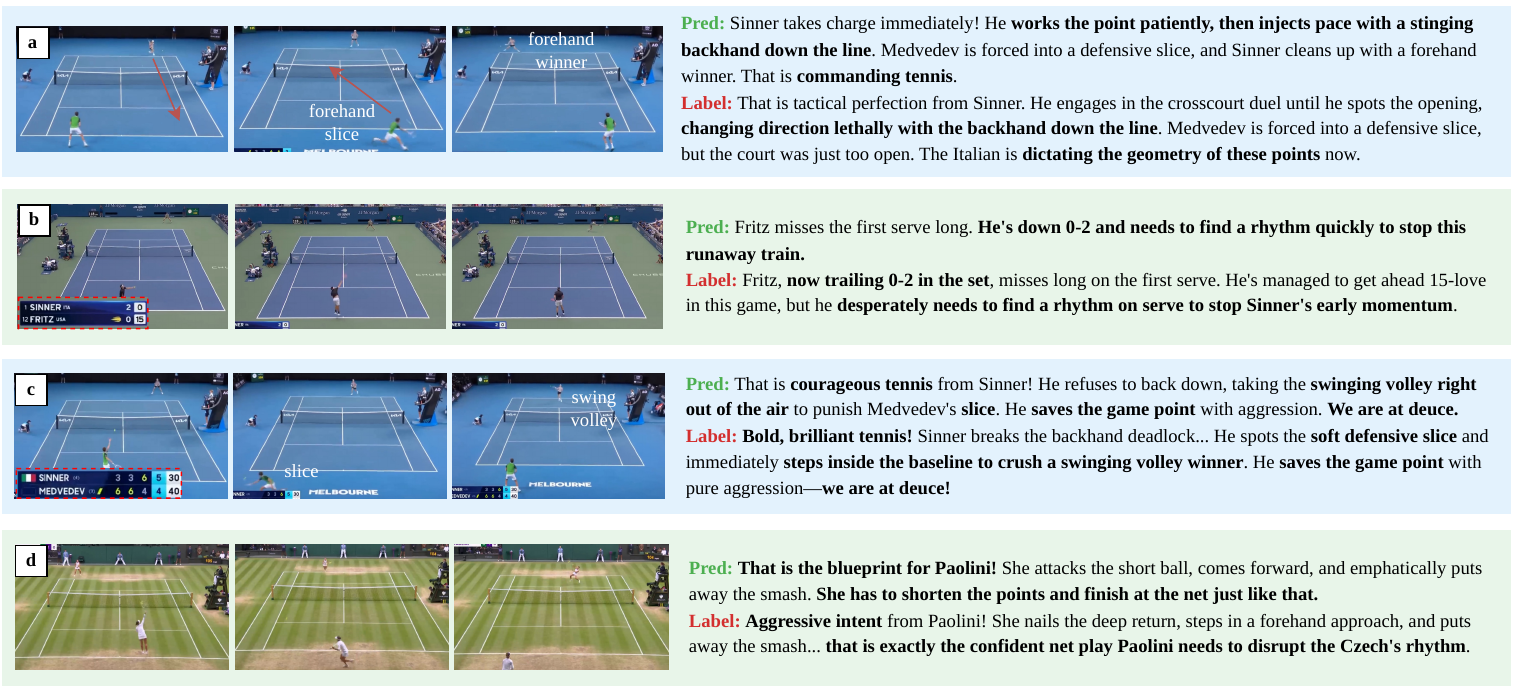}
    \caption{\textbf{Qualitative comparison of commentary generation.} Our method produces expert-level commentary across multiple dimensions: \textbf{(a)} tactical depth, \textbf{(b)} temporal momentum, \textbf{(c)} professional terminology, and \textbf{(d)} strategic prediction.}
    \label{fig:qualitative_commentary}
    \vspace{-15px}
\end{figure*}

To demonstrate the efficacy of our framework, we visualize the intermediate outputs of our video semantic parser and compare the predicted generations against the ground truth.
Figure~\ref{fig:vid_parser} shows qualitative results of the video semantic parser across hard, clay, and grass courts. It reliably detects court corners, players, and ball locations under varying court appearances and broadcast viewpoints.

Figure~\ref{fig:qualitative_commentary} presents representative rally-level examples with keyframes, model predictions, and ground truth. 
The generated commentaries demonstrate strong alignment with expert annotations, capturing tactical depth, temporal momentum, and score awareness. For instance, the model describes point construction (e.g., building pressure before striking down the line), recognizes critical score contexts (e.g., ``down 0–2'' or ``game point at deuce''), and employs professional terminology such as ``swinging volley'' and ``dictating geometry''. These examples illustrate the model’s ability to contextualize individual shots within broader match dynamics and produce analytically rich, expert-level commentary.

\vspace{-6px}
\section{Conclusion}
\vspace{-6px}
We present \dataset{}, the first large-scale benchmark for expert-level tennis commentary generation, and \model{}, a fine-grained, context-aware framework for tennis understanding. \dataset{} contains 470+ hours of broadcast video and 40,000+ rally clips with analytical commentary. Built on a video semantic parser and hierarchical memory, \model{} captures both rally dynamics and match context. Experiments show that it outperforms leading open-source and proprietary MLLMs while remaining efficient for real-time deployment.

\clearpage
%
%
\bibliographystyle{splncs04}
\bibliography{main}

\clearpage
\appendix
\section{Automated Rally Segmentation Details}
\label{sec:appendix_segmentation}

In this section, we provide additional details about the automated rally segmentation process used during dataset construction.

Professional tennis broadcasts typically consist of alternating periods of rally play and pauses between points. To organize the dataset at the rally level, we segment full match broadcasts into individual rally clips. 

To achieve this efficiently across hundreds of hours of video, we employ a lightweight, purely audio-based detection model. Specifically, we train a neural network to identify the distinct acoustic signatures of racquet-ball impacts from the broadcast audio track. To ensure the model is robust against complex ambient stadium noises, we augment our training data by overlaying various background sounds (e.g., crowd cheering, clapping, and announcer voices) onto the audio samples. This augmentation strategy allows the model to reliably distinguish true hitting events from similar impulsive noises in the broadcast.

Once the hitting moments are identified, these temporal anchors are clustered in the temporal domain to form candidate rally intervals. We then apply specific filtering rules to retain valid segments corresponding to complete, uninterrupted rallies. In particular, we retain clips that satisfy the following conditions:

\begin{itemize}
\item The clip contains a standard broadcast top-down court view.
\item The scoreboard is clearly visible within the frame.
\end{itemize}

These criteria ensure that each extracted clip provides sufficient and consistent contextual information for downstream multimodal analysis.

\section{LLM Prompt for Commentary Generation}
\label{sec:appendix_prompt_commentary}

To construct high-quality analytical commentary for each rally, we employ a large language model to synthesize commentary from multimodal match metadata. In our pipeline, we use the Gemini~3~Pro API in a chat-based interaction mode.

\subsection{Prompt Design}

The model is instructed to assume the role of a professional tennis expert. The prompt encourages the model to focus on tactical reasoning, match momentum, and strategic interpretation rather than simple play-by-play narration.

The system prompt used in our implementation is shown below.

\begin{promptbox}{System Prompt}
I want you to act as a professional tennis commentator and coach. I will give you descriptions of tennis matches in progress, which include both detailed shot-by-shot data and real broadcast transcripts. You will commentate on the match, providing your analysis on what has happened thus far and predicting how the match will go.

You should be knowledgeable of tennis terminology, tactics, players involved in each match. Your commentary must be factually accurate based on the shot data. Explicitly use the broadcast transcripts to extract long-term match context, tactical shifts, and any rolling match statistics mentioned by the original commentators (e.g., serve percentages, error counts). Weave these macro trends into your commentary naturally when they add strategic depth to the current point.

Be professionally insightful, engaging the audience, and maintain narrative coherence by connecting the current rally to the momentum of the recent points and the overall match story. Ensure the length is appropriate and use natural pauses (ellipses...) and transitions. Focus on intelligent analysis rather than just narrating play-by-play.
\end{promptbox}

The user prompt provides structured metadata for multiple consecutive rallies. In our implementation, the model receives metadata for 10 chronological rallies at a time. The model is asked to generate a commentary for each rally.

\begin{promptbox}{User Prompt}
Here is the metadata for N chronological rallies. Provide a commentary for each rally between 5 and 60 words, adjusting the length based on rally duration and importance. Return the output strictly as a valid JSON list of strings.

Metadata: ...
\end{promptbox}

\subsection{Input Metadata Structure}

For each rally, the prompt includes structured metadata describing the match context, rally sequence, and outcome. The metadata is organized in JSON format. The main components of the metadata include:

\begin{itemize}
\item \textbf{match\_info}: Tournament, round, court surface, and player identities.
\item \textbf{score\_state (initial)}: Current scoreboard state at the beginning of the rally, including set score, game score, point score, and server identity.
\item \textbf{rally}: Shot-by-shot sequence describing the rally progression.
\item \textbf{outcome}: Final result of the rally, including the point winner and reason for the point ending.
\item \textbf{audio\_transcription}: Broadcast transcript text aligned with the rally clip, which provides additional contextual cues about match momentum and commentary context.
\end{itemize}

A simplified example of the metadata format is shown below.

\begin{promptbox}{Metadata}
{
  ``match_info'': {
    ``tournament'': ``...'',
    ``round'': ``...'',
    ``surface'': ``...'',
    ``player_1'': {``name'': ``...'', ``handedness'': ``...''},
    ``player_2'': {``name'': ``...'', ``handedness'': ``...''}
  },

  ``score_state (initial)'': {
    ``server'': ``...'',
    ``returner'': ``...'',
    ``sets'': {...},
    ``games_in_current_set'': {...},
    ``points_in_current_game'': {...}
  },

  ``rally'': [
    {
      ``shot_index'': 0,
      ``hitter'': ``...'',
      ``shot_description'': ``...''
    }
  ],

  ``outcome'': {
    ``point_winner'': ``...'',
    ``point_loser'': ``...'',
    ``reason'': ``...''
  },

  ``audio_transcription (background context)'': ``...''
}
\end{promptbox}

\subsection{Conversation Context Handling}
The Gemini API is used in chat mode. To maintain short-term contextual continuity across consecutive prompts while preventing excessive context growth, we keep only the most recent interaction in the chat history.
This design allows the model to maintain limited conversational continuity while keeping the token length manageable during large-scale annotation.

\section{Prompt Design for Scoreboard Understanding}
\label{sec:scoreboard_prompt}

To accurately extract the match state without processing redundant visual information, we crop the broadcast frame using the bounding box of the scoreboard area and provide only this region as the visual input to the MLLM. 

Since the visual layout, server indicators, and logical structure of scoreboards vary significantly across different Grand Slam tournaments, we design tournament-specific prompts. This tailored approach helps the MLLM accurately parse complex structures and handle edge cases, such as the Advantage (``AD'') scoring or missing point columns.

\subsection{Australian Open and US Open}
For the Australian Open and US Open, the scoreboards share a similar standard layout where the player names are in the leftmost column, followed by a server icon, and then the sequence of game and point scores. 

\begin{promptbox}{Prompt for Australian Open \& US Open}
You are given an image of a tennis scoreboard.

### Scoreboard Structure
The structure of the scoreboard is based on the following rules:
- Each column has two rows, which are strictly vertically aligned.
- The **leftmost column** displays the names of two players, each on a separate row.
- The **second column** indicates the current server:
  - The currently-serving player will have an **icon** displayed in this column.
  - The non-serving player's will have this column left **empty**.
- **From left to right**, the **subsequent columns** represents either the **game score** within a set or the **point value** within a game of each player. 

### Special cases:
- When one row in the last column has a value ``AD'', the other row in this column is left empty in the image.
  - In this case, you must fill the empty row with value ``40''. 
  - Always extract value ``40'' from the empty row.   
  - Do not mix an ``AD'' value in the last column with values in previous columns.
  - Do not guess or hallucinate an ``AD'' case - extract from the image only.

Think step by step. Your output should include:
- Each player's **game score** within a set or the **point value** within a game, which are stored in a list. These two lists must have the same length.
- The **current server**
- Make sure each row has the same number of columns. 
- Do not miss any columns that you see in the image. 
- Do not hallucinate any columns - extract from the image only.

### Output Format:
{
    ``PLAYER1_NAME'': [``column_1'', ``column_2'', ...],
    ``PLAYER2_NAME'': [``column_1'', ``column_2'', ...],
    ``server'': ``PLAYER1_NAME'' or ``PLAYER2_NAME''
}

### Example:
Scoreboard:
[Name]   [Serve]   [Column1]  [Column2] [Column3] [Column4]   
Alice                  6          1         1                             
Bob        icon        4          6         2         AD
  
Expected Output:
{
  ``Alice'': [``6'', ``1'', ``1'', ``40''],
  ``Bob'': [``4'', ``6'', ``2'', ``AD''],
  ``server'': ``Bob''
}

Only return valid JSON (no explanation, no markdown, no code blocks). This output should be directly parsable by Python's `json.loads()`.
\end{promptbox}

\subsection{Roland Garros (French Open)}
The Roland Garros scoreboard typically places the server indicator (often a slash ``/'' or ``//'') in the very first column, preceding the player names. The prompt is adjusted to capture this specific trait.

\begin{promptbox}{Prompt for Roland Garros}
You are given an image of a tennis scoreboard (typical of French Open/Roland Garros style).

### Scoreboard Structure
The structure of the scoreboard is based on the following rules:
- Each column has two rows, which are strictly vertically aligned.
- The **leftmost column** (Column 1) indicates the current server:
  - The currently-serving player will have an indicator (usually ``/'' or ``//'') 
    in this column.
  - The non-serving player will have this column left **empty**.
- The **second column** displays the names of the two players.
- **From left to right**, the **subsequent columns** represents either the 
  **game score** within a set or the **point value** within a game of each player. 

### Special cases:
- When one row in the last column has a value ``AD'', the other row in this column 
  is left empty in the image.
  - In this case, you must fill the empty row with value ``40''. 
  - Always extract value ``40'' from the empty row.   
  - Do not mix an ``AD'' value in the last column with values in previous columns.
  - Do not guess or hallucinate an ``AD'' case - extract from the image only.

Think step by step. Your output should include:
- Each player's scores (sets, games, and points) stored in a list. 
  These two lists must have the same length.
- The **current server** (identified by the indicator in the first column).
- Make sure each row has the same number of columns. 
- Do not miss any columns that you see in the image. 
- Do not hallucinate any columns - extract from the image only.

### Output Format:
{
    ``PLAYER1_NAME'': [``column_1'', ``column_2'', ...],
    ``PLAYER2_NAME'': [``column_1'', ``column_2'', ...],
    ``server'': ``PLAYER1_NAME'' or ``PLAYER2_NAME''
}

### Example:
Scoreboard:
[Serve]   [Name]     [Set1]  [Set2] [Point]   
  /       Alice        6       1      40                             
          Bob          4       6      AD

Expected Output:
{
  ``Alice'': [``6'', ``1'', ``40''],
  ``Bob'': [``4'', ``6'', ``AD''],
  ``server'': ``Alice''
}

Only return valid JSON (no explanation, no markdown, no code blocks). 
This output should be directly parsable by Python's `json.loads()`.
\end{promptbox}

\subsection{Wimbledon}
Wimbledon scoreboards often employ a distinct structure where the points column disappears entirely between games, and the server is denoted by a triangle icon. We explicitly instruct the model to handle the missing point column dynamically.

\begin{promptbox}{Prompt for Wimbledon}
You are given an image of a tennis scoreboard (typical of Wimbledon style).

### Scoreboard Structure
The structure of the scoreboard is based on the following rules:
- Each column has two rows, which are strictly vertically aligned.
- The **leftmost column** displays the names of two players.
- The **second column** (immediately to the right of the name) indicates 
  the current server:
  - The currently-serving player will have a **triangle icon** (often purple, 
    pointing left <) displayed next to their name.
  - The non-serving player will have this space left **empty**.
- The **subsequent columns** (from left to right) follow this specific order:
  1. **Set Score**: The column (often with a green background) showing the 
     number of sets won.
  2. **Game Score**: The column (often with a purple background) showing the 
     current games in the active set.
  3. **Point Value**: The last column on the right (often with a white background). 
     **Note:** This column may be completely absent in the image (e.g., between games).

### Special Handling Rules
1. **Missing Points Column**: 
   - Look for the white column to the right of the purple ``Game Score'' column.
   - If this column is **not visible** (the scoreboard ends at the purple column), 
     you must set the **Point Value** to **``0''** for both players.
   
2. **``AD'' Scoring (Only if Points Column is visible)**:
   - If one row in the Points column has the value ``AD'', the other row in this 
     column is often left empty.
   - In this specific case, fill the empty row with value ``40''. 

Think step by step. Your output should include:
- Each player's scores stored in a list. The list must always contain 3 elements: 
  `[Set Score, Game Score, Point Score]`.
- The **current server**.

### Output Format:
{
    ``PLAYER1_NAME'': [``set_score'', ``game_score'', ``point_score''],
    ``PLAYER2_NAME'': [``set_score'', ``game_score'', ``point_score''],
    ``server'': ``PLAYER1_NAME'' or ``PLAYER2_NAME''
}

### Example 1 (Standard - Points Visible):
Scoreboard Image shows: [Name] [Serve] [Sets] [Games] [Points]
Alice < 1 2 15
Bob     1 2 30

Output:
{
  ``Alice'': [``1'', ``2'', ``15''],
  ``Bob'': [``1'', ``2'', ``30''],
  ``server'': ``Alice''
}

### Example 2 (Hidden Points Column):
Scoreboard Image shows: [Name] [Serve] [Sets] [Games] (Points column is missing)
Alice < 0 2
Bob     1 2

Output:
{
  ``Alice'': [``0'', ``2'', ``0''],
  ``Bob'': [``1'', ``2'', ``0''],
  ``server'': ``Alice''
}

Only return valid JSON (no explanation, no markdown, no code blocks). 
This output should be directly parsable by Python's `json.loads()`.
\end{promptbox}

\section{Additional Implementation Details}

\subsection{Video Semantic Parser}

For scoreboard understanding, we localize and crop scoreboard regions and apply OCR to the center frame of each rally clip for score recognition.

Event spotting is trained end-to-end by downscaling clips to $224 \times 224$, extracting frame-wise features using a video encoder, and applying a dense prediction head to infer fine-grained event sequences and timestamps~\cite{liu2025f}.

Player and ball coordinates are obtained using an RF-DETR~\cite{sapkota2025rf} detector fine-tuned on tennis data. Court corners are detected via heatmap-based prediction of 14 keypoints followed by homography estimation, allowing projection of ball landing points and player positions into real-world court coordinates.

\subsection{Short-Term Memory Construction}

We adopt a sliding-window short-term memory with $K=4$, retaining the current rally and the previous four rallies.

To avoid redundant video encoding across overlapping windows, all video segments in a conversation are pre-encoded once at the beginning of each training step. The resulting visual embeddings are cached and reused during window construction. Corresponding visual tokens are sliced from the cache rather than recomputed.

Each sliding window forms an independent rally-level training unit. Loss is computed per window and backpropagated immediately. This chunk-wise backward strategy prevents computation graphs from spanning multiple windows and significantly reduces GPU memory usage.

To avoid duplicated supervision caused by overlapping windows, tokens outside the newly introduced rally segment are masked. Each token contributes to the loss exactly once.

\subsection{Supervised Fine-Tuning Setup}

We adopt \textit{Qwen3-VL-8B-Instruct} as the backbone. Supervised fine-tuning (SFT) updates the language model parameters while freezing both the vision encoder and the multimodal projector to preserve pretrained visual representations.

Training is implemented using \textit{LLaMA-Factory} on 4 NVIDIA H200 GPUs with BF16 mixed precision. The model is trained for 3 epochs with a learning rate of $1\times10^{-5}$, cosine annealing scheduling, and 10\% warmup. The per-device batch size is 1 with 2 gradient accumulation steps.

\subsection{Distributed Training Stability}

Under distributed data parallel (DDP) training, conversations may produce different numbers of sliding windows. Unequal backward passes across ranks can lead to communication deadlocks.

To ensure synchronization, we compute the global minimum number of windows across ranks via distributed reduction. Each rank randomly samples its local windows (without replacement) to match this minimum. A shared random seed is broadcast to guarantee consistent collective participation. This ensures aligned backward passes and stable multi-GPU training.

\section{LLM-based Evaluation Prompt}
\label{sec:appendix_prompt_evaluation}

To better assess the quality of generated tennis commentary, we introduce an LLM-based evaluation protocol in addition to standard text generation metrics.

In this protocol, an LLM evaluator acts as a senior tennis analyst and compares the predicted commentary with the ground-truth reference. The evaluator assigns scores based on five criteria: ``Accuracy'', ``Coherence'', ``Excitement'', ``Professionalism'', and ``Pacing''.

Each criterion measures a different aspect of commentary quality, including factual correctness, narrative structure, tactical depth, and stylistic appropriateness. The system and user prompts used for this evaluation are provided below.

\begin{promptbox}{System Prompt for Evaluation}
You are a senior Tennis Analyst and expert commentator evaluator. Your task is to evaluate a **Generated Commentary** against strict **Match Metadata** and a **Reference Commentary** (Ground Truth).
\end{promptbox}

\begin{promptbox}{User Prompt for Evaluation}
### INPUT DATA:
1. **METADATA (Ground Truth):** {metadata}
2. **REFERENCE COMMENTARY (Style/Tone Baseline):** ``{reference}''
3. **PREDICTION (Target to Evaluate):** ``{prediction}''

### SCORING RUBRIC (0-20 points per category, Total 100):
1. **ACCURACY (0-20 pts):** Alignment with METADATA (players, shot types, score, court positions).
   - 20: Perfect factual match.
   - 0: Hallucinations (wrong player, wrong shot) or contradictions with Metadata.
2. **COHERENCE (0-20 pts):** Logical flow and pronoun usage.
   - 20: Natural narrative; events connect logically.
   - 0: Confusing structure; contradictions within the text.
3. **EXCITEMENT (0-20 pts):** Tone matches the event intensity.
   - 20: Highly engaging; emotive vocabulary fitting the moment.
   - 0: Robotic, flat, or mismatched tone (e.g., boring description of a winner).
4. **PROFESSIONALISM (0-20 pts):** Domain terminology and depth of analysis.
   - 20: Insightful observation (e.g., noting ``inside-out forehand'' or ``tactical adjustment'').
   - 0: Superficial or generic description only.
5. **PACING (0-20 pts):** Length relative to event complexity.
   - 20: Concise for quick points; descriptive for long rallies.
   - 0: Severe mismatch (e.g., long paragraph for a simple double fault).

### OUTPUT INSTRUCTION:
Provide your evaluation **strictly** as a Python dictionary string (JSON compatible). 
Do NOT output any markdown or conversational text. 
The dictionary must have the following keys:
{
    ``scores'': {
        ``accuracy'': <int>,
        ``coherence'': <int>,
        ``excitement'': <int>,
        ``professionalism'': <int>,
        ``pacing'': <int>
    },
    ``total_score'': <int>
}
\end{promptbox}

\section{Additional Qualitative Examples}
\label{sec:appendix_qualitative}

\begin{figure*}[h]
    \centering
    \includegraphics[width=\textwidth]{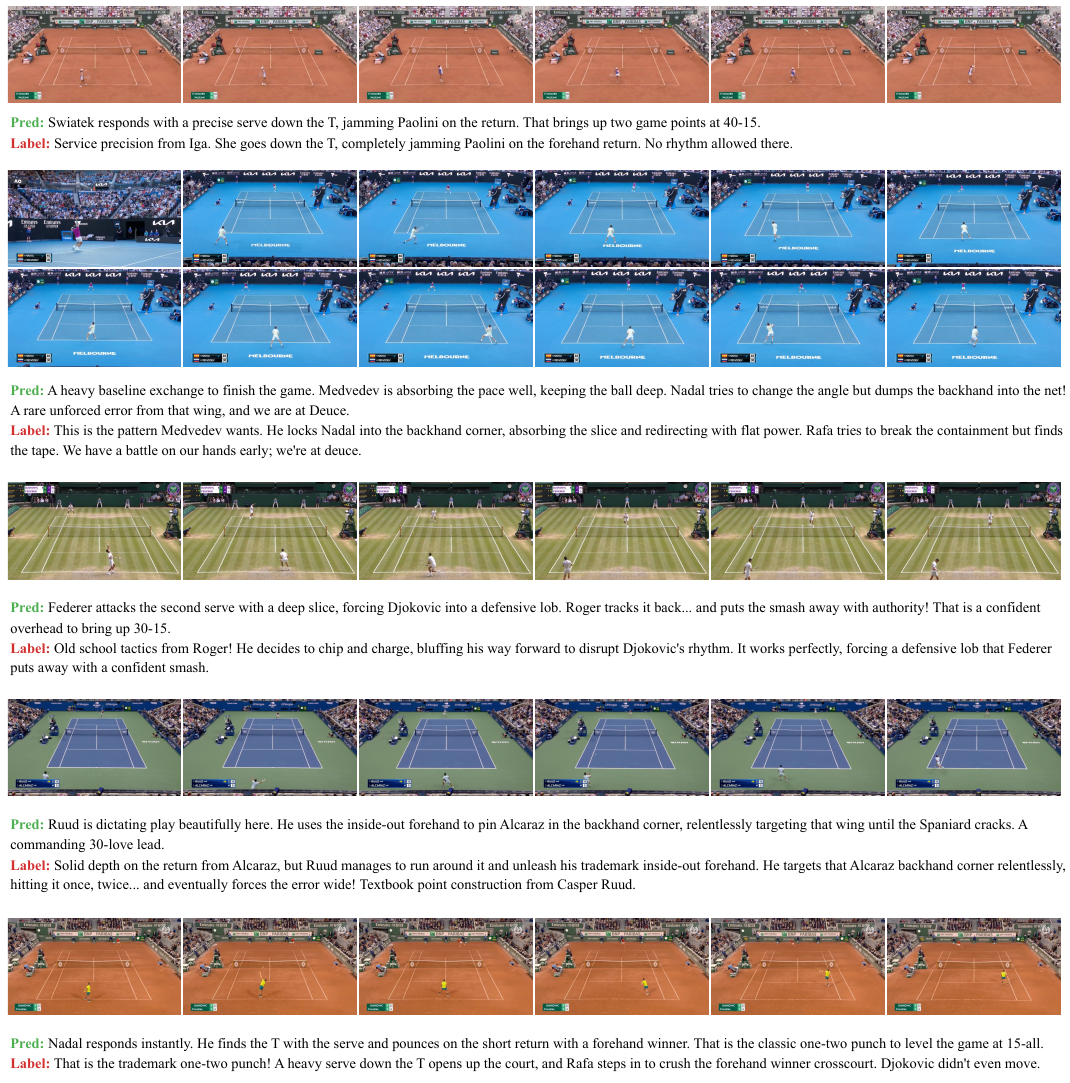}
    \caption{More qualitative results on commentary generation.}
    \label{fig:more_qualitative}
    \vspace{-5px}
\end{figure*}

We provide additional qualitative examples (Figure~\ref{fig:more_qualitative}) to further illustrate the capabilities of the proposed system. These examples highlight different match scenarios, including rallies with varied tactical patterns, score contexts, and shot sequences.
Each example includes representative video frames together with the predicted commentary and corresponding reference annotations.

\section{Ethical Considerations for \dataset{}} \label{sec:appendix_ethics}

The \dataset{} dataset is compiled entirely from publicly accessible broadcast videos sourced from YouTube. This section details the ethical frameworks guiding our data collection, copyright adherence, privacy protection, and efforts to minimize bias.

To strictly comply with copyright laws and platform terms of service, we do not distribute raw video files. Instead, \dataset{} provides YouTube URLs paired with our annotations. This ensures that the original content creators and rightsholders retain full control over their media. 

Privacy concerns are mitigated by the nature of the data: the dataset focuses exclusively on professional athletes competing in highly publicized, internationally broadcasted events. 
\dataset{} does not introduce any additional personal information beyond what is already publicly available in the broadcast footage. We mandate that this dataset be utilized strictly for academic research in sports video understanding and automated commentary generation, explicitly prohibiting its use for commercial or non-research applications.

To promote fairness and mitigate bias, \dataset{} was designed to be highly diverse. The collection spans all four major Grand Slam tournaments and includes a wide array of professional players from both the ATP and WTA tours, covering various court surfaces and handedness profiles. We do not filter subjects based on race, nationality, or gender. However, we encourage the research community to remain vigilant, evaluate models for emerging biases during training, and provide feedback so that future iterations of the dataset remain equitable and representative.

In summary, \dataset{} is carefully constructed to balance the advancement of automated sports analytics with rigorous ethical standards. By relying on public broadcasts, limiting usage to academic research, and safeguarding player privacy, we offer a robust benchmark that respects the rights of all stakeholders.

\end{document}